\documentclass[10pt,twocolumn,letterpaper]{article}

\usepackage[pagenumbers]{cvpr} 

\definecolor{cvprblue}{rgb}{0.21,0.49,0.74}
\usepackage[pagebackref,breaklinks,colorlinks,allcolors=cvprblue]{hyperref}
\usepackage{algorithm}
\usepackage{algorithmic}
\usepackage{xcolor}
\usepackage{graphicx}
\usepackage{amsmath}
\usepackage{capt-of}
\usepackage{adjustbox}
\usepackage{cuted}
\usepackage{xcolor} 
\usepackage{colortbl} 
\usepackage{graphicx} 
\usepackage{booktabs}
\usepackage{color}
\usepackage[utf8]{inputenc}
\usepackage{array}
\usepackage{multirow}
\usepackage{amssymb} 
\usepackage{tabularx}
\usepackage{tikz}
\usepackage{kotex}

\newcommand{\figref}[1]{Fig.~\ref{#1}}
\newcommand{\tabref}[1]{Tab.~\ref{#1}}
\newcommand{\secref}[1]{Sec.~\ref{#1}}

\newcommand*\samethanks[1][\value{footnote}]{\footnotemark[#1]}

\newcommand{\tikzxmark}{%
\tikz[scale=0.23] {
    \draw[line width=0.7,line cap=round] (0,0) to [bend left=6] (1,1);
    \draw[line width=0.7,line cap=round] (0.2,0.95) to [bend right=3] (0.8,0.05);
}}
\newcommand{\tikzcmark}{%
\tikz[scale=0.23] {
    \draw[line width=0.7,line cap=round] (0.25,0) to [bend left=10] (1,1);
    \draw[line width=0.8,line cap=round] (0,0.35) to [bend right=1] (0.23,0);
}}

\title{A Training-Free Style-Personalization via SVD-Based Feature Decomposition}

\author{Kyoungmin Lee\footnote{}~, Jihun Park\samethanks~, Jongmin Gim\samethanks~,\\ Wonhyeok Choi, Kyumin Hwang, Jaeyeul Kim and Sunghoon Im\footnote{}~~\\
DGIST, Daegu, Republic of Korea\\
{\tt\small \{kyoungmin, pjh2857, jongmin4422, smu06117, kyumin, jykim94, sunghoonim\}@dgist.ac.kr}
}

\begin{document}
\maketitle
\begin{strip} 
    \centering
    \includegraphics[width=.88 \linewidth]{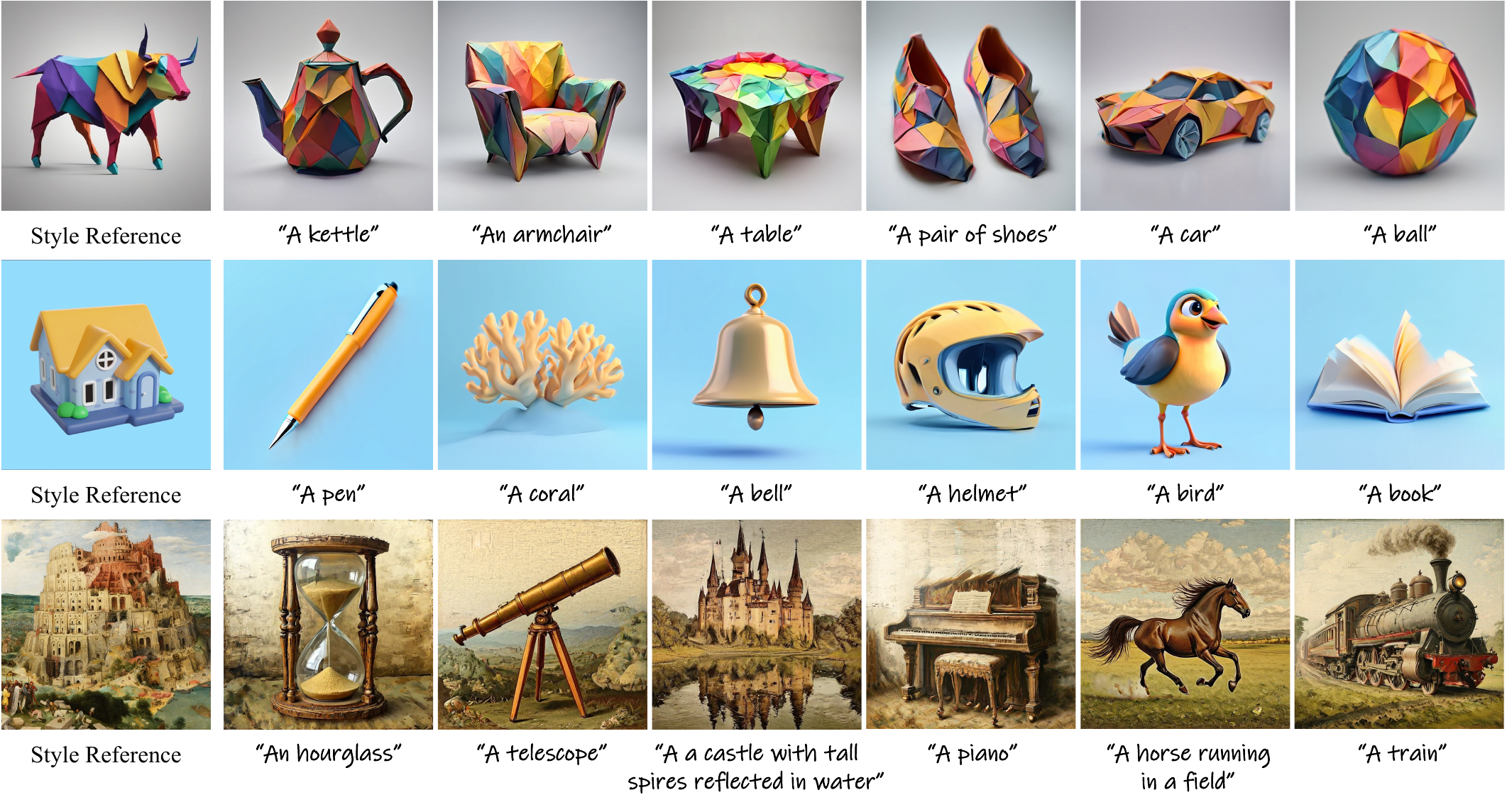}
    \captionof{figure}{Style-personalized image generation results produced by our method. Given reference style images and text prompts, our method generates images with consistent style and diverse content.}

    \label{fig:teaser}
\end{strip}

\footnotetext[1]{Equal contribution.}
\footnotetext[2]{Corresponding author.}

\renewcommand*{\thefootnote}{\arabic{footnote}}

\begin{abstract}
We present a training-free framework for style-personalized image generation that operates during inference using a scale-wise autoregressive model. 
Our method generates a stylized image guided by a single reference style while preserving semantic consistency and mitigating content leakage.
Through a detailed step-wise analysis of the generation process, we identify a pivotal step where the dominant singular values of the internal feature encode style-related components.
Building upon this insight, we introduce two lightweight control modules: \textit{Principal Feature Blending}, which enables precise modulation of style through SVD-based feature reconstruction, and \textit{Structural Attention Correction}, which stabilizes structural consistency by leveraging content-guided attention correction across fine stages.
Without any additional training, extensive experiments demonstrate that our method achieves competitive style fidelity and prompt fidelity compared to fine-tuned baselines, while offering faster inference and greater deployment flexibility.
\end{abstract}
    
\section{Introduction}
\label{sec:intro}

Text-to-Image (T2I) models~\cite{ramesh2021zero, rombach2022high, saharia2022photorealistic, podell2023sdxl, chang2023muse, han2025infinity, tang2024hartefficientvisualgeneration} have rapidly transformed the creative landscape, enabling artists, designers, and casual users alike to generate high-quality visuals from natural language prompts. Fueled by massive image-text datasets~\cite{changpinyo2021conceptual, lin2014microsoft, schuhmann2022laion, kakaobrain2022coyo-700m}, these models now support an unprecedented range of content diversity and stylistic expression.
As creative tools become increasingly democratized, users are seeking more than just visually plausible outputs—they desire personalized generation that reflects specific visual identities~\cite{ruiz2023dreambooth, li2023blip, wei2023elite, ryu2023low, li2024photomaker} or preferred artistic styles~\cite{sohn2023styledrop, hertz2024stylealignedimagegeneration, liu2024unziplora, ryu2023low, zhang2024finestyle}. These emerging demands call for generation systems that are not only high-quality but also customizable, efficient, and responsive to individual preferences.

Existing solutions~\cite{shah2023ziplorasubjectstyleeffectively, liu2024unziplora, ahn2024dreamstyler, frenkel2024implicitstylecontentseparationusing} have made progress in this direction, often relying on fine-tuning mechanisms to encode style-specific characteristics.
However, such methods typically involve training a new model instance per style, posing scalability challenges in real-world applications.
Additionally, most systems are built on diffusion-based T2I architectures~\cite{podell2023sdxl, saharia2022photorealistic, flux2024}. Although these model produce high-quality results, their iterative denoising process leads to slow inference, which makes them less suitable for real-time or interactive applications.

Motivated by the limitations, we propose a novel style-personalized image generation framework that combines efficiency, flexibility, and stylistic fidelity. 
Our method generates high-quality images guided by a single reference style image during inference without any additional training. 
To achieve this, we leverage a large-scale text-to-image model, specifically a scale-wise autoregressive model \cite{han2025infinity}, which offers significantly faster inference compared to diffusion models while maintaining strong visual fidelity. 

To better understand and maximize the capabilities of this scale-wise autoregressive model, we conduct a detailed analysis of its generation process.
Our analysis reveals that a specific step in the generation process plays a crucial role in determining both content and style. 
At this step, the dominant singular values of the feature play a key role, as they effectively capture and separate the style-related components.
Building on this insight, we develop a \textit{Principal Feature Blending} mechanism that enables precise control over style, leveraging a specific step feature and allowing the model to faithfully reflect the stylistic traits inherent in the style features of a reference image.
Additionally, we introduce a \textit{Structural Attention Correction} strategy, which stabilizes the generation process by leveraging content-related information to preserve structural consistency.
By integrating these components, our training-free framework achieves high-quality image generation, as illustrated in \figref{fig:teaser}. 
It also achieves competitive performance in both quantitative and qualitative evaluations, while maintaining significantly faster inference time.

In summary, our contributions include:
\begin{itemize}
    \item We present a training-free inference framework for style-personalized image generation from a single style reference, achieving competitive results with significantly faster inference.
    \item We conduct a detailed step-wise analysis of the scale-wise generation process and identify a key step that governs both content and style.
    \item We observe that style-related components can be effectively extracted from the feature at this step through an SVD-based analysis of its dominant singular values.
    \item We propose two lightweight control modules—\textit{Principal Feature Blending} for precise style modulation and \textit{Structural Attention Correction} for stabilizing structural coherence in generation.
\end{itemize}
\section{Related Work}

\subsection{Neural Style Transfer}
Image stylization, which alters the visual style of an image, has become an active area of research. A major breakthrough came with Neural Style Transfer (NST) \cite{gatys2016image}, which used a pre-trained CNN (VGGNet) to separately extract content and style features. While effective, NST required costly per-image optimization. To tackle this issue, \cite{huang2017arbitrary} proposed Adaptive Instance Normalization (AdaIN), aligning the mean and variance of the content features to those of the style features for faster style transfer. Building upon this, subsequent works such as \cite{li2017universal, lu2019closed} introduced the Whitening and Coloring Transform (WCT), which aligns the full covariance structure of the features, resulting in more detailed and higher-quality stylization.
With the rise of attention mechanisms in neural networks \cite{vaswani2017attention, dosovitskiy2020image}, recent models further advanced stylization quality by utilizing attention to achieve remarkable stylization results \cite{liu2021adaattn, hong2023aespa, deng2020arbitrary, park2019arbitrary, yao2019attention, gim2024content}. 
Parallel to this, vision-language models such as CLIP \cite{radford2021learning} have enabled text-driven style transfer \cite{park2025styleeditortextdrivenobjectcentricstyle, kwon2022clipstyler,bar2022text2live}, allowing intuitive control via natural language, without requiring explicit style images. This expands the scope of style transfer beyond traditional image-based paradigms.

\subsection{Text-to-Image Generation}

Recent advances in large-scale image-text datasets \cite{changpinyo2021conceptual, lin2014microsoft, schuhmann2022laion, kakaobrain2022coyo-700m} have greatly enhanced the ability of models to bridge the gap between natural language and visual modalities, fueling progress in conditional image synthesis.
This has spurred the development of large-scale Text-to-Image (T2I) generation frameworks—including diffusion-based models \cite{ramesh2021zero, rombach2022high, saharia2022photorealistic, podell2023sdxl, flux2024}, GAN-based approaches \cite{kang2023scaling}, and visual autoregressive (AR) models \cite{chang2023muse, han2025infinity, tang2024hartefficientvisualgeneration}—which can generate diverse, high-quality images from natural language prompts.
Diffusion-based models have become the dominant T2I paradigm due to their superior image quality and success in downstream tasks like style transfer and image editing \cite{chung2024style, yang2023zero, tumanyan2022plugandplay, brooks2023instructpix2pix, jeong2024trainingfree, kwon2023diffusionbased, hertz2023deltadenoisingscore, poole2023dreamfusion}.
However, their high inference latency poses challenges for real-time applications.
Meanwhile, visual AR models have evolved from traditional next-token prediction \cite{van2017neural, esser2021taming} to more efficient masked token prediction \cite{chang2022maskgit, chang2023muse, kondratyuk2023videopoet}.
The recent introduction of the next-scale prediction paradigm \cite{tian2024visual} has further accelerated inference without compromising output quality, establishing next-scale AR models \cite{han2025infinity, tang2024hartefficientvisualgeneration, voronov2025swittidesigningscalewisetransformers} as promising alternatives to diffusion-based methods.

\begin{figure*}[t]
    \centering
    \includegraphics[width=.9\linewidth]{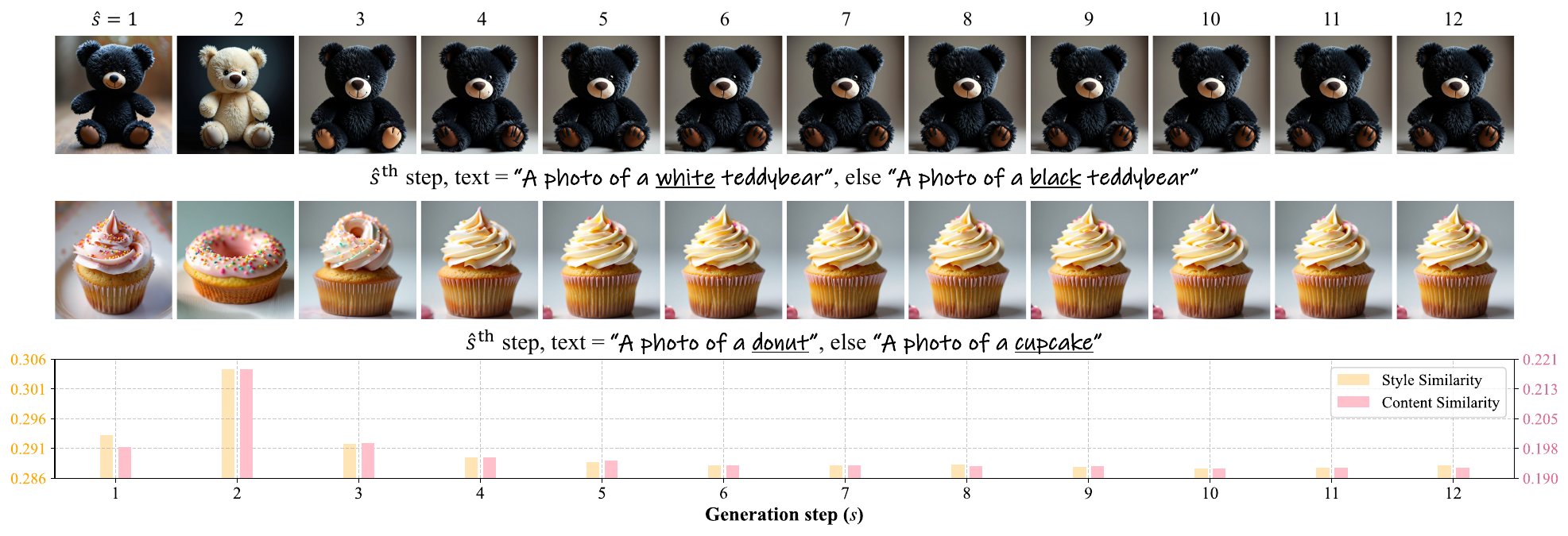}
    \caption{Step-wise prompt injection analysis.
We intervene at each generation step \(s \in \{1, \dots, 12\}\) by replacing the prompt only at step \(\hat{s}\), while keeping all other steps fixed to the base prompt. 
\textbf{Top}: style prompt injection (``\texttt{A photo of a black teddy bear}'' vs. ``\texttt{A photo of a white teddy bear}'').  
\textbf{Middle}: content prompt injection (``\texttt{A photo of a cupcake}'' vs. ``\texttt{A photo of a donut}'').  
\textbf{Bottom}: CLIP similarity between the alternative prompt \({\hat{T}}\) and the corresponding image across steps. 
}
\label{fig:observation}
\end{figure*}

\subsection{Personalized image generation}
Recent advances in personalized image generation have led to methods that adapt novel visual concepts to user intent using pre-trained T2I models. These methods are generally categorized into content-oriented and style-oriented approaches. Content-oriented methods \cite{ruiz2023dreambooth, li2023blip, wei2023elite, li2024photomaker} aim to capture object-specific or identity-preserving features from a small set of user-provided reference images. By fine-tuning pre-trained models or injecting learned embeddings, they generate images that maintain a high degree of fidelity to the target subject.
Building on the technical foundations of these content-oriented personalization methods, recent work has extended similar principles to style-oriented personalized generation \cite{sohn2023styledrop, frenkel2024implicitstylecontentseparationusing, shah2023ziplorasubjectstyleeffectively, hertz2024stylealignedimagegeneration, ryu2023low, zhang2024finestyle, ahn2024dreamstyler, park2025training}. In this paradigm, the objective shifts from preserving content identity to consistently controlling the visual style across diverse generations.
Despite their effectiveness, these methods predominantly rely on diffusion-based models and often require fine-tuning, which introduces high computational costs and long inference times. In contrast, we propose a training-free, scale-wise autoregressive model that achieves fast style-personalized image generation based on a comprehensive analysis of the scale-wise autoregressive model.

\section{Preliminary}
\label{sec: preliminary}

\paragraph{Infinity Architecture.}
In our work, we leverage Infinity~\cite{han2025infinity}, a state-of-the-art T2I framework that employs the next-scale prediction paradigm introduced by \cite{tian2024visual} to generate high-fidelity, text-aligned images.
During inference time, the Infinity architecture is composed of three key components: a pre-trained text encoder $\mathcal{E}_T$ based on Flan-T5 \cite{chung2022scalinginstructionfinetunedlanguagemodels}, an autoregressive transformer $\mathcal{M}$ that performs scale-wise feature prediction, and a decoder $\mathcal{D}$ that reconstructs the final image from accumulated residual feature maps.

At each generation step $s \in \mathbf{S}$, where $\mathbf{S} = \{1, 2, \dots, S\}$ denotes the set of all generation steps, the autoregressive transformer $\mathcal{M}$ iteratively predicts a $s$-th scale quantized residual feature map $R_s$, conditioned on the input text prompt ${T}$ and the previously generated feature $F_{s-1}$.
The process begins with initial features ${F}_0$ corresponding to the start-of-sequence \( \langle \text{SOS} \rangle \) token.
The prediction process is defined as:
\begin{gather}
\begin{split}
\label{eq:residual}
{R}_s &= \mathcal{M}({F}_{s-1}, \mathcal{E}_T({T})) \\
&= M_{CA}(M_{SA}({Q}_{s-1}, {K}_{s-1}, {V}_{s-1}), \mathcal{E}_T({T})),~\forall s \in \mathbf{S},
\end{split}
\end{gather}
where ${Q_s}$, ${K_s}$, and ${V_s}$ are the query, key, and value at $s$-th generation step projected from feature $F_s$ respectively.
Here, $M_{SA}(\cdot)$ and $M_{CA}(\cdot)$ denote the self-attention and cross-attention mechanisms within the transformer.

Each predicted residual ${R}_s$ is upsampled to the resolution $H \times W$ using a bilinear upsampling function $\text{up}_{H\times W}(\cdot)$, and the resulting features are accumulated across scales to form the next-step input:
\begin{equation}
{F}_s = \sum_{i=1}^{s} \text{up}_{H\times W}(\mathbf{R}_i), \quad {R}_s \in \mathbb{R}^{c \times h_s \times w_s},
\end{equation}
where $h_s$ and $w_s$ denote the spatial dimensions of the residual features at step $s$, $c$ denote the channel of the quantized feature.
The final image ${I}$ is produced by decoding the accumulated representation ${F}_S$ at the final generation step:
\begin{equation}
{I} = \mathcal{D}({F}_S).
\end{equation}

\begin{figure}[t]
    \centering
    \includegraphics[width=.92\linewidth]{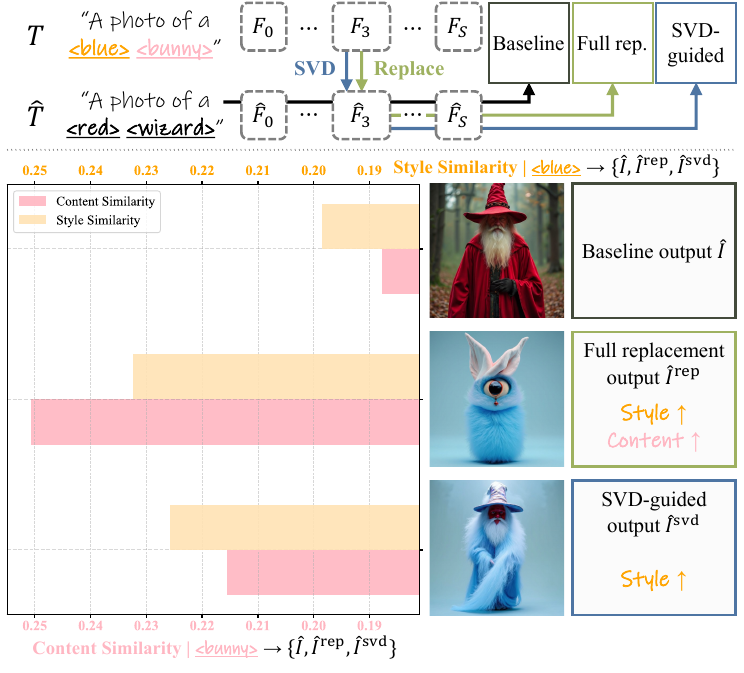}
    \caption{Key step feature analysis. Content and style similarity are measured for Baseline, Full replacement, and SVD-guided outputs using a set of prompt pairs $\mathbf{T}$, with results averaged across all pairs.}
    \label{fig:observation_svd}
    \vspace{-12pt}
\end{figure}

\begin{figure*}[t!]
    \centering
    \includegraphics[width=0.82\textwidth]{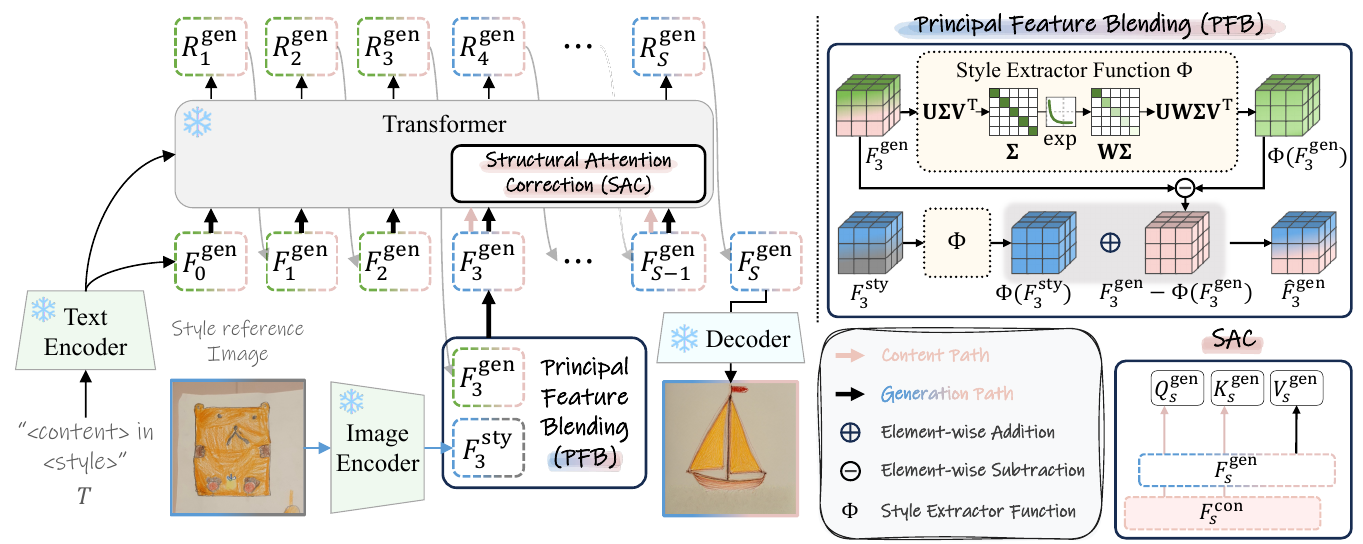}
    \captionof{figure}{\textbf{Overall pipeline of our model}. 
    The text encoder processes an identical text prompt \(T\) for both the content and generation paths, providing their embeddings to the autoregressive transformer. At stage $s=3$, \textit{Principal Feature Blend} is applied to extract the principal style representation from the reference style image and seamlessly integrate it into the features of the generation path. Starting from $s=3$ (the fine stage), \textit{Structural Attention Correction} aligns the generation path’s attention maps with those of the content path, ensuring stable and consistent structural guidance during refinement.}
    \label{fig:overall}
\end{figure*}

\section{Analysis of Scale-wise AR Model}
\label{sec:analysis}

\paragraph{(1) Step-wise analysis.}
We investigate the internal mechanisms of each step in the scale-wise autoregressive model's generation process, focusing on its influence over two key visual attributes of the generated image: \textit{content} and \textit{style} representations.
To facilitate this analysis, we construct two prompt pair sets $(T, \hat{T})$ $\in$ $\mathbf{T}^{\text{con}} \cup \mathbf{T}^{\text{sty}}$ from 100 base prompts, initially generated by ChatGPT:
\begin{itemize}
    \item \textbf{Content pair set} $\mathbf{T}^{\text{con}}$: This set contains 100 prompt pairs $(T^\text{con}$, $ \hat{T}^\text{con}) \in \mathbf{T}^\text{con}$, each formed by randomly selecting two distinct object-centric base prompts (e.g., ``\texttt{A photo of a donut}'', ``\texttt{A photo of a truck}'').
    \item \textbf{Style pair set} $\mathbf{T}^{\text{sty}}$: For each of the 100 base object prompts, a style pair $(T^\text{sty}, \hat{T}^\text{sty}) \in \mathbf{T}^\text{sty}$ was created by assigning two different colors (from a set of 10 predefined colors) to the same object, while keeping object category fixed (e.g., ``\texttt{A photo of a red truck}'' and ``\texttt{A photo of a green truck}'').
\end{itemize}

Then, for both $\textbf{T}^\text{con}$ and $\textbf{T}^\text{sty}$, we generate modified images $\hat{I}$, as shown in \figref{fig:observation}, by replacing the original text prompt $T$ with an alternative prompt $\hat{T}$ at a specific generation step $s \in \mathbf{S}$ in Eq.~\ref{eq:residual}, while keeping all other components unchanged.
A substantial change in the resulting image in response to this step-specific prompt injection indicates that the corresponding step plays a critical role in shaping certain visual attributes.

To assess the impact of each generation step, we compute the CLIP similarity \cite{radford2021learning} between alternative prompt \({\hat{T}}\) and its corresponding generated image as shown in the bottom row in \figref{fig:observation}. 
We observe that step $\hat{s} = 2$ consistently produces the highest similarity with $\hat{T}$ across all 200 prompt pairs.
This result suggests that step 2 plays a key role in shaping both content and style attributes. Consequently, the \textbf{third feature} $F_3$, generated after this step, plays a crucial role in determining the final output image.

\paragraph{(2) Key step feature analysis.}
As discussed in the previous section, the third feature $F_3$ plays a pivotal role in shaping both content and style. Building on this observation, and supported by prior studies \cite{park2025training,nguyen2025csd} showing that early-step features in scale-wise autoregressive models often encode strong stylistic cues, we hypothesize that the principal components of $F_3$ are predominantly shaped by stylistic attributes.

To analysis this, we first construct a set of 100 prompt pairs $\mathbf{T}$, where each pair $(T, \hat{T})$ differs in both object category and color (e.g., ``\texttt{A photo of a red truck}'' and ``\texttt{A photo of a purple cat}''). 
For each prompt $T$, we apply singular value decomposition (SVD) to the third feature, yielding $F_3 = U \Sigma V^\top$. We then construct a modified diagonal matrix $\Sigma'$, obtained by zeroing out all singular values except the largest one $\sigma_1$. Using this matrix, we reconstruct the dominant singular component as $F_3^{\text{svd}} = U \Sigma' V^\top$. The corresponding residual is defined as $F_3^{\text{res}} = F_3 - F_3^{\text{svd}}$. We perform the same decomposition for the feature $\hat{F}_3$ obtained from the prompt $\hat{T}$.
For each prompt pair $(T, \hat{T})$, we evaluate the effect of manipulating the dominant singular component by generating three outputs:
\begin{itemize}
    \item \textbf{Baseline output $\hat{I}$}, generated using the original prompt $\hat{T}$ without any feature manipulation.
    \item \textbf{Full replacement output $\hat{I}^{\text{rep}}$}, obtained by directly substituting the entire feature $\hat{F}_3$ with ${F}_3$. ($\hat{F}_3 \leftarrow {F}_3$)
    \item \textbf{SVD-guided output $\hat{I}^{\text{svd}}$}, obtained by replacing only the dominant singular component of $\hat{F}_3$ with that of ${F}_3$, while preserving the residual component. ($\hat{F}_3 \leftarrow {F}_3^{\text{svd}}+\hat{F}_3^{\text{res}}$)
\end{itemize}

As shown in \figref{fig:observation_svd}, the full-replacement output $\hat{I}^{\text{rep}}$ displays a substantial increase in CLIP similarity to both the object (e.g., ``\texttt{bunny}'') and the color (e.g., ``\texttt{blue}'') described in the substituted prompt ${T}$. In contrast, the SVD-guided output $\hat{I}^{\text{svd}}$ shows a pronounced increase primarily in color-related CLIP similarity, while changes in object-related similarity remain much smaller. This result suggests that modifying only the dominant singular component primarily affects stylistic attributes, with minimal influence on content. Together, these observations suggest that the first principal component of the third feature $F_3$ predominantly captures style-related characteristics, with limited contribution from content-related information.

\section{Method}
\subsection{Overall pipeline}

In this paper, we aim to generate a style-personalized image \(I^\text{gen}\) by injecting a principal style feature into the final generation image, while preserving semantic consistency and suppressing content leakage. 

As illustrated in \figref{fig:overall}, our method employs a dual-stream generation architecture composed of a \textit{content path} and a \textit{generation path}, both conditioned on the same text prompt \(T\) (``\texttt{<content> in <style>}''). Using an identical prompt prevents semantic mismatch between the two streams and enables consistent structural communication during inference. The \textit{content path} operates as the standard inference branch of the pre-trained model without any modification and follows the iterative update rule in \eqref{eq:residual}, producing a sequence of content features $\{F_s^{\text{con}}\}_{s=1}^{S}$. Its role is to provide structurally stable and semantically aligned guidance throughout the generation process. In parallel, the \textit{generation path} follows the same update formulation but produces its own feature sequence $\{F_s^{\text{gen}}\}_{s=1}^{S}$, which is subsequently modulated by our proposed style-blending mechanism. This path synthesizes the final stylized output while leveraging structural cues from the content path and incorporating style information in a controlled and targeted manner.

Building upon this dual-stream iterative process, we introduce two complementary modules—\textit{Principal Feature Blending} (PFB) and \textit{Structural Attention Correction} (SAC)—which operate exclusively on the generation path while leveraging cues from the other streams. PFB (\secref{sec:attnshare}) selectively injects principal style representations from the style reference image into the generation features, with a targeted intervention at the third step ($s=3$) to prevent style-unrelated feature leakage. 
Following this style injection, SAC (\secref{sec:query}) is applied across the subsequent steps, where it incorporates \textit{content path} signals to stabilize structural alignment and maintain semantic consistency throughout the refinement process.

\subsection{Principal Feature Blending}
\label{sec:attnshare}

Our process begins by extracting the style features \({F}_{s}^\text{sty}\) using a pretrained multi-scale image encoder $\mathcal{E}_I$ from the baseline Infinity \cite{han2025infinity}, as follows:
\begin{equation}
    \{{F}_{1}^\text{sty}, {F}_{2}^\text{sty}, \dots, F_S^\text{sty}\} = \mathcal{E}_I(I^\text{sty}).
\end{equation}
Among these multi-scale features, the step-wise analysis in \secref{sec:analysis} identifies the third feature \({F}_{3}^\text{sty}\) as a pivotal representation that strongly influences both content and style during generation. 
Motivated by this observation, we focus on \({F}_{3}^\text{sty}\) as the primary carrier of style information and use it as the basis for our style modulation mechanism.

To effectively incorporate style information while suppressing irrelevant cues from the reference image, we introduce \textit{Principal Feature Blending}, a mechanism that selectively transfers the principal components of the style feature into the generation process. 
This design is grounded in two analyses from \secref{sec:analysis}:  
(1) the \textbf{Step-wise analysis}, which pinpoints \({F}_{3}\) as the most influential feature for stylistic control, and  
(2) the \textbf{Key step feature analysis}, which reveals that the dominant singular values of \({F}_3^\text{sty}\) encode the core stylistic characteristics.  
Guided by these insights, our method extracts the principal style representation from \({F}_3^\text{sty}\) and blends it into the generation path with minimal disruption to the original content structure.

To achieve seamless blending in the feature space, we design a style extractor function $\Phi$, which prioritizes the dominant contribution of the leading component while smoothly incorporating residual style representations. 
Based on the observation that the first singular value of $F_3^\text{sty}$ acts as the primary carrier of style information, we enhance its influence while retaining minor contributions from the remaining components to preserve stylistic continuity. 
Accordingly, $\Phi$ applies exponential reweighting to the singular values based on their spectral order, gradually reducing the impact of lower components:
\begin{equation}
\begin{gathered}
    \Phi(F) \triangleq \mathbf{U W \Sigma V^\top}, \\
    \text{where~~} F = \mathbf{U \Sigma V^\top}, \\
    \mathbf{W} = \text{diag}(\exp^{-0 \cdot \alpha}, \exp^{-1 \cdot \alpha}, \dots, \exp^{-(r-1) \cdot \alpha}),
\end{gathered}
\end{equation}
where $r$ denotes the rank of the feature matrix, and $\alpha>0$ controls the exponential decay rate along the singular spectrum. 

To substitute the generation path’s style representation with the one extracted from the style feature, we incorporate the refined style via $\Phi$ and update the generation feature:
\begin{equation}
\begin{gathered}
    F_3^{\text{gen}} \leftarrow \hat{F}_3^{\text{gen}},\\
    \hat{F}_3^{\text{gen}} =
    \Phi({F}_3^{\text{sty}})
    + \big({F}_3^{\text{gen}} - \Phi({F}_3^{\text{gen}})\big).
\end{gathered}
\end{equation}
This formulation preserves the original structure information of the generation path while seamlessly injecting style information derived from the reference.

\subsection{Structural Attention Correction}
\label{sec:query}
While PFB effectively integrates style cues into the generation path, we observed that its feature-level modulation can unintentionally disturb the structural coherence of generated results, sometimes causing spatial misalignment or shape distortion. To stabilize the generation process, we leverage the attention map of the content path as a structural prior, inspired by the self-attention mechanism in diffusion-based architectures \cite{shin2025exploringmmdit, tumanyan2023plug}, where the interaction between Queries and Keys preserves spatial and structural relationships. 
Building on this, we introduce \textit{Structural Attention Correction (SAC)}, which aligns the attention map of the generation path with that of the content path to ensure consistent structural guidance throughout the generation process. 

SAC is applied to all subsequent steps following the application of Principal Feature Blending (PFB), denoted as $\mathbf{S}_\text{fine} = \{3,4, \dots, S\}$. 
These steps correspond to the stages where content and style representations continue to interact. Formally, SAC injects the content queries and keys at each step $s \in \mathbf{S}_\text{fine}$ as follows:

\begin{equation}
\begin{gathered}
    Q_s^\text{gen} \leftarrow Q_s^\text{con}, \quad K_s^\text{gen} \leftarrow K_s^\text{con},\\
    Q_s^{\text{con}} = {W}_Q F_s^{\text{con}},\quad K_s^{\text{con}} = {W}_K F_s^{\text{con}}, \\[4pt]
\end{gathered}
\end{equation}
Here, \(W_Q\) and \(W_K\) denote the linear projection matrices that transform input features into query and key representations in the self-attention layers. \(Q_s^\text{con}\) and \(Q_s^\text{gen}\) denote the content and generation queries at step \(s\), and \(K_s^\text{con}\) and \(K_s^\text{gen}\) denote the corresponding keys.

\section{Experiments}

\begin{table*}[t]
\caption{
Quantitative comparison with state-of-the-art style-personalized image generation models. 
The symbols $\uparrow$ and $\downarrow$ indicate that higher and lower values are better, respectively. The inference time is measured as the time required to generate a single image. For tuning-based methods (\textit{StyleDrop}, \textit{DreamStyler}, \textit{DB-LoRA}, and \textit{B-LoRA}), we present the combined inference time, which accounts for both the tuning phase (given the reference style image) and the inference time required to produce a single output.
}
\centering
\footnotesize
\resizebox{0.95\linewidth}{!}{

\begin{tabular}
{c|c|cccccccc}
\toprule
Metric  & \textbf{Ours} & IP-Adapter  & StyleAligned  & StyleDrop & DreamStyler & DB-LoRA & B-LoRA & CSGO & StyleAR\\
\midrule
Training-Free & \tikzcmark & \tikzxmark & \tikzcmark & \tikzxmark & \tikzxmark & \tikzxmark & \tikzxmark & \tikzxmark & \tikzxmark\\ 
\midrule
Harmonic score (\(S_{\text{harmonic}}\)) $\uparrow$ & \underline{0.437} & 0.433 & \textbf{0.438} & 0.386 & 0.403 & 0.420 & {0.410} & 0.421 & 0.434\\
Prompt fidelity (\(S_{\text{txt}}\)) $\uparrow$ & \textbf{0.334} & 0.302 & 0.315 & 0.273 & 0.304 & 0.323 & \underline{0.324} & 0.318 & 0.314\\
Style fidelity (\(S_{\text{img}}\)) $\uparrow$ & 0.630 & \textbf{0.763} & \underline{0.716} & 0.657 & 0.599 & 0.602 & 0.559 & 0.623 & 0.701 \\
\midrule
Inference time (seconds) $\downarrow$ & \textbf{3.58} & \underline{10.13} & 64.58 & 520.07 & 698.98 & 342.01 & 630.42 & 15.87 & 346.68\\
\bottomrule
\end{tabular}
}
\label{tab:comparison}
\end{table*}

\begin{figure*}[t]
    \centering
    \includegraphics[width=.92\linewidth]{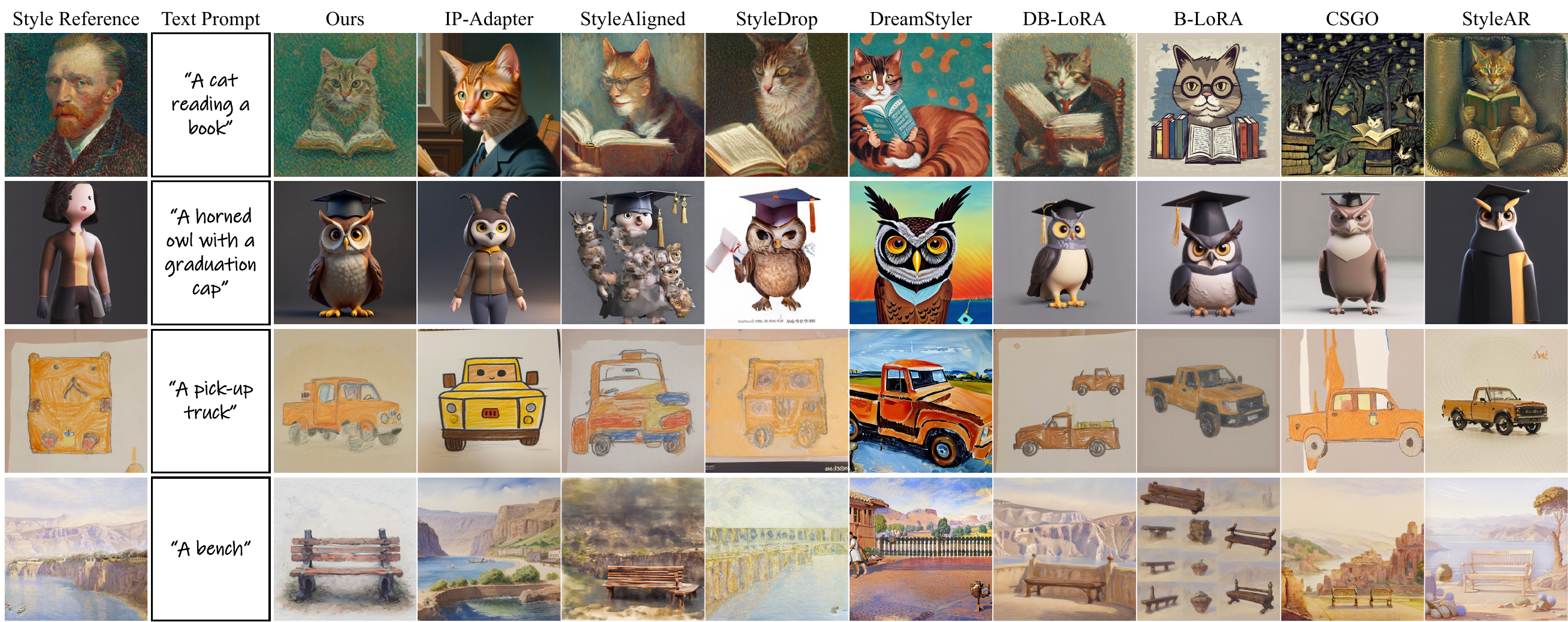}
    \caption{Qualitative comparison with state-of-the-art style-personalized image generation models.}
    \label{fig:qualitative_comparison}
\end{figure*}

\subsection{Implementation Details}

We implement our method using a pre-trained Infinity 2B model~\cite{han2025infinity} with all parameters frozen, which performs scale-wise prediction across 12 steps. The baseline employs a codebook of size $2^{32}$, with quantized feature maps of resolution $64\times 64\times 32$. The exponential decay rate $\alpha$ for Principal Feature Blending is set to 1.0.

Based on our analysis, our method operates in a step-wise manner with targeted interventions across the generation process. At $s=3$, we apply \textit{Principal Feature Blending}, 
and at the fine stages ($\mathbf{S}_\text{fine} = \{3, 4, \dots, S\}$), we apply \textit{Structural Attention Correction}. Generating a 1024$\times$1024 style-personalized image takes approximately 3.58 seconds on a single NVIDIA A6000 GPU.

\subsection{Evaluation Setup}
\label{sec:evaluation}
\paragraph{Benchmark.}
We follow the evaluation protocol introduced in FineStyle~\cite{zhang2024finestyle}, synthesizing images by combining a filtered subset of prompts from Parti~\cite{yu2022scaling} with 10 representative styles from the evaluation set (see Appendix for details).
The Parti subset consists of 190 prompts, each describing a subject along with its superclass to reduce semantic ambiguity (e.g., A cat, animals, in watercolor painting style).

\paragraph{Evaluation Metrics.}
Following FineStyle, we evaluate generated images using two CLIP-based metrics: 
$S_{\text{txt}}$ (CLIP Text score) and $S_{\text{img}}$ (CLIP Image score).
$S_{\text{txt}}$ measures the similarity between each generated image and its corresponding text prompt to assess prompt fidelity, while $S_{\text{img}}$ measures the similarity between the generated image and a reference style image to assess style fidelity. 
However, a higher $S_{\text{img}}$ does not always imply better stylization quality, 
as excessive similarity may result from content leakage or mode collapse, as mentioned in FineStyle. To provide a more balanced evaluation, we additionally report the harmonic mean of the two scores, denoted as $S_{\text{harmonic}}$ (Harmonic score), which jointly reflects both prompt and style fidelity. Formally, it is computed as:
\begin{equation}
S_{\text{harmonic}} = \frac{2  S_{\text{txt}} S_{\text{img}}}{S_{\text{txt}} + S_{\text{img}}}.    
\end{equation}

\subsection{Comparison with state-of-the-art style-personalized image generation models}

To demonstrate the performance and efficiency of our model, we compare our method against eight state-of-the-art style-personalized image generation models: StyleDrop \cite{sohn2023styledrop}, StyleAligned \cite{hertz2024stylealignedimagegeneration}, IP-Adapter \cite{ye2023ipadaptertextcompatibleimage}, DreamStyler \cite{ahn2024dreamstyler}, DreamBooth-LoRA (DB-LoRA) \cite{ryu2023low}, B-LoRA \cite{frenkel2024implicitstylecontentseparationusing}, CSGO \cite{xing2024csgo}, StyleAR \cite{wu2025stylear}. 

In \tabref{tab:comparison}, we quantitatively compare our method with state-of-the-art baselines. While StyleAligned and IP-Adapter show relatively high style fidelity ($S_{\text{img}}$), they exhibit noticeably lower prompt fidelity ($S_{\text{txt}}$), indicating limited semantic alignment with the input text. As highlighted in \cite{zhang2024finestyle}, high $S_{\text{img}}$ scores can be misleading due to issues like content leakage or mode collapse, where the model mimics the reference style image instead of transferring style.
This effect is evident in the qualitative results shown in \figref{fig:qualitative_comparison}, where models with high $S_{\text{img}}$ scores, such as StyleAligned and IP-Adapter, frequently exhibit content leakage (first and second row). In these cases, structural details from the style reference are unintentionally transferred into the output image, leading to degraded content fidelity. These findings emphasize that high style similarity alone is insufficient to ensure faithful and semantically aligned image generation.

In contrast, DB-LoRA and B-LoRA achieve relatively high scores in prompt fidelity $S_\text{txt}$. However, they require additional fine-tuning for each new style reference, which limits scalability in practical applications. Moreover, all training-based methods suffer from long inference times, ranging from tens to hundreds of seconds per image, due to the overhead of iterative denoising or fine-tuning. Our method, by contrast, is fully training-free, up to 195$\times$ faster, and achieves competitive results, making it well-suited for real-time and interactive use cases.
As illustrated in \figref{fig:qualitative_comparison}, DB-LoRA and B-LoRA tend to better preserve the semantics of the input prompt, reinforcing that $S_{\text{txt}}$ is a reliable indicator of semantic alignment, even when $S_{\text{img}}$ alone may be misleading. However, despite their strong prompt adherence, both methods tend to show relatively weaker style fidelity compared to ours, suggesting that the reference style may be only partially reflected in some cases. In contrast, our method reliably preserves both the intended content and the reference style. Despite being the fastest among all methods, it still achieves a strong balance between prompt fidelity and style fidelity, underscoring its practical advantage for high-quality, real-time style-personalized generation.

\subsection{Ablation study}
\label{sec:ablation}

The quantitative results in \tabref{tab:ablation} highlight how each proposed component contributes to achieving a balance between style fidelity and prompt fidelity. To clearly demonstrate the effect of our PFB module, we compare two variants: a \textbf{direct feature replacement} strategy (REP) and our \textbf{Principal Feature Blending} (PFB).  As shown in \tabref{tab:ablation}-(a), the baseline configuration attains the highest prompt fidelity ($S_\text{txt}$) but exhibits limited style fidelity due to the absence of explicit style modulation. In contrast, directly replacing the style feature in (b) yields the highest style fidelity ($S_\text{img}$), but at the expense of severe prompt degradation, indicating significant content leakage from the style reference. The SVD-based blending in (c) provides a more favorable trade-off: it mitigates the prompt-fidelity drop observed in (b) while still offering a substantial improvement in style fidelity, consistent with our observation that the dominant singular component primarily captures stylistic information. Finally, the full model in (d), which integrates both PFB and SAC, achieves the most balanced performance across all metrics, yielding the highest harmonic score ($S_\text{harmonic}$). This demonstrates that the proposed modules effectively complement one another, enhancing style fidelity with minimal sacrifice of prompt fidelity.

\begin{figure}[t]
    \centering
    \includegraphics[width=.9\linewidth]{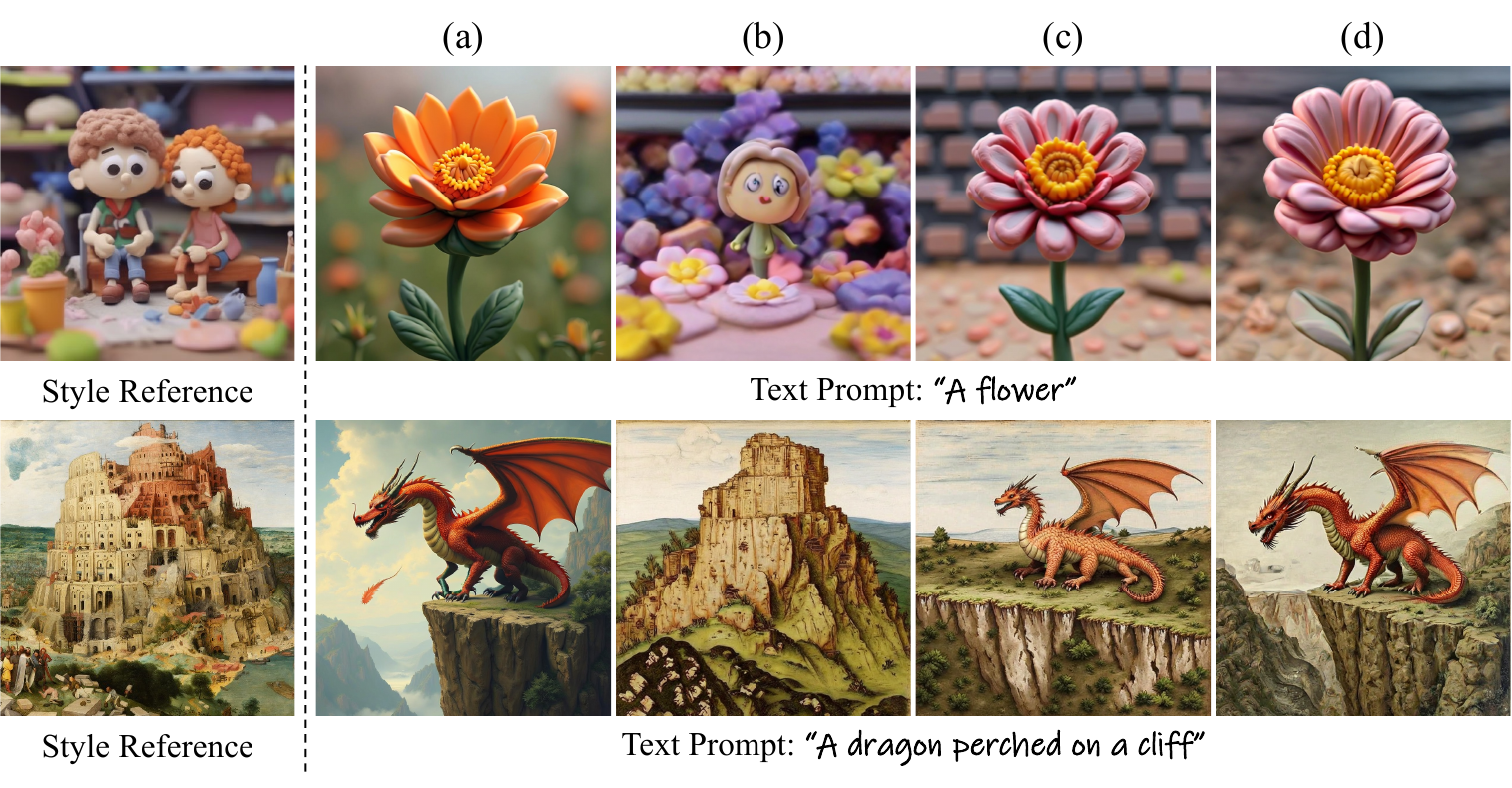}
    \caption{Qualitative ablation study on proposed method. (a)-(d) correspond to the component in \tabref{tab:ablation}.}
    \label{fig:ablation}
\end{figure}

The qualitative comparison in \figref{fig:ablation} further supports these trends. In \figref{fig:ablation}-(a), while the baseline produces clean and coherent images, it fails to reproduce the stylistic characteristics of the reference. Direct replacement in (b) enforces strong style transfer but also introduces unintended content elements from the reference, resulting in clear prompt mismatch. The SVD-guided variant in (c) successfully captures the intended style while retaining the target content, though its prompt adherence is still weaker than the baseline. In contrast, the full model in (d) preserves the style of the reference and simultaneously generates images that align closely with the prompt, achieving the most balanced and desirable output—consistent with the quantitative trends observed above.

\begin{table}[t]
    \caption{
    Ablation study on Principal Feature Blending (PFB) and Structural Attention Correction (SAC). {REP} denotes replacement using the style feature ${F}_3^{\text{sty}}$. The symbol $\uparrow$ indicates that higher is better. The best and second-best results are highlighted in \textbf{bold} and \underline{underline}, respectively.
    }
    \centering
    \small
    \resizebox{.9\linewidth}{!}{
    \begin{tabular}{c|l|ccc} 
    \toprule
      \#  & Method  
      & $S_\text{txt}$ $\uparrow$ 
      & $S_\text{img}$ $\uparrow$ 
      & $S_\text{harmonic}$ $\uparrow$ \\
    \midrule
     (a) & Infinity
         & \textbf{0.348} & 0.559 & \underline{0.429} \\
     (b) & Infinity + REP 
         & 0.279 & \textbf{0.696} & 0.398 \\
     (c) & Infinity + PFB 
         & 0.321 & \underline{0.631} & 0.426 \\
     (d) & Infinity + PFB + SAC
         & \underline{0.334} & 0.630 & \textbf{0.437} \\
    \bottomrule
    \end{tabular}
    }
    \label{tab:ablation}
\end{table}

\subsection{User study}
We conduct a user study with 30 participants (ages 20s--50s) to further support our evaluation. Participants evaluate two key aspects: prompt and style fidelity. 
We selected comparison models based on their quantitative performance: StyleAligned~\cite{hertz2024stylealignedimagegeneration} and IP-Adapter~\cite{ye2023ipadaptertextcompatibleimage}, 
which achieved the highest $S_{\text{img}}$ (style fidelity) scores, 
and DB-LoRA~\cite{ryu2023low} and B-LoRA \cite{frenkel2024implicitstylecontentseparationusing}, which achieved the highest $S_{\text{txt}}$ (prompt fidelity) scores. 
Our method achieves a {clearly superior preference in prompt fidelity} (35.3\%) while maintaining {competitive style fidelity} (32.0\%), compared to the other models' scores of 4.3\%, 5.0\%, 26.7\%, 28.7\% (prompt) and 30.7\%, 23.3\%, 8.3\%, 5.7\% (style). An example of the interface is in the Supplementary material.

\section{Conclusion}
In this work, we introduced a training-free framework for style-personalized image generation that operates on a single reference image and leverages the efficiency of a scale-wise autoregressive model. Through a detailed step-wise analysis of the model’s generation process, we identified a pivotal feature that jointly governs content and style, and further demonstrated—via an SVD-based spectral study—that its dominant singular component captures style-specific variation. Building on these insights, we proposed two lightweight yet effective modules: \textit{Principal Feature Blending}, which provides precise and interpretable style control, and \textit{Structural Attention Correction}, which stabilizes structural consistency during generation.
Our method achieves high performance while preserving prompt fidelity, offering a favorable balance. Quantitative and qualitative evaluations confirm that the proposed components operate as intended, enabling faithful style personalization without additional training and with significantly faster inference than existing models.

\clearpage
\appendix
\maketitlesupplementary

\section{Comprehensive analysis of our method}

\subsection{Additional Results for Key Step Feature Analysis}

In \secref{sec:analysis}-(2), we showed that replacing only the largest singular component of $F_3$ primarily alters stylistic attributes while largely preserving content. To further validate this observation, we extend the SVD-guided manipulation experiment by varying the number of preserved singular values.

We use the same prompt setup and intervention protocol as in the main paper: we construct 100 mixed prompt pairs $(T, \hat{T})$, each differing in both object category and color (e.g., ``\texttt{A photo of a red truck}'' vs.\ ``\texttt{A photo of a purple cat}''). For each prompt, we perform singular value decomposition $F_3 = U \Sigma V^\top$, and reconstruct truncated variants that retain only the top-$k$ components:
\begin{equation}
    F_3^{(k)} = U \Sigma^{(k)} V^\top,
\end{equation}
where $\Sigma^{(k)}$ is constructed by preserving only the largest $k$ singular values while zeroing out the remaining entries. We evaluate $k \in \{1, 2, 4, 8, 16, 32\}$, and for each $k$, we generate SVD-guided outputs by replacing the corresponding portion of $\hat{F}_3$:
\begin{equation}
    \hat{F}_3 \leftarrow F_3^{(k)} + \hat{F}_3^{\text{res}(k)},
\end{equation}
where $\hat{F}_3^{\text{res}(k)} = \hat{F}_3 - \hat{F}_3^{(k)}$ preserves the remaining feature components.

\begin{figure}[h]
    \centering
    \includegraphics[width=.95\linewidth]{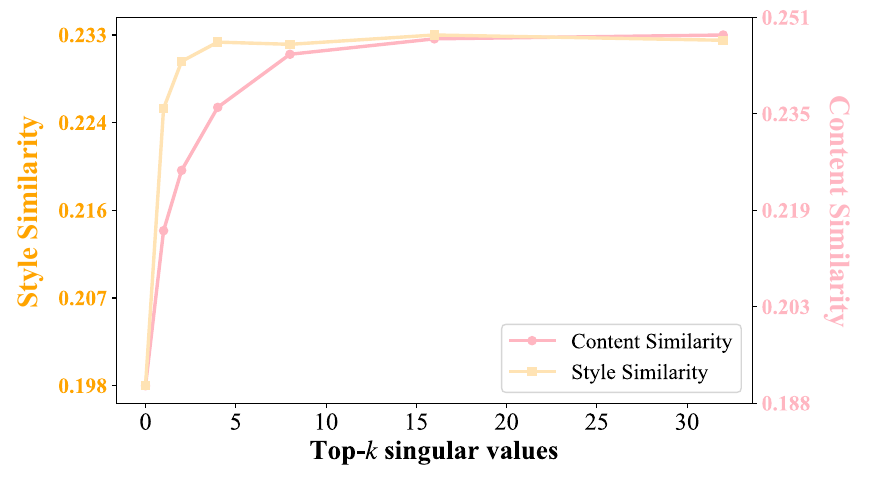}
    \caption{Qualitative results of SVD-guided feature replacement with varying top-$k$ singular values. From left to right: the baseline output generated from $\hat{T}$, SVD-guided outputs with $k \in \{1,2,4,8,16,32\}$, and the baseline output generated from $T$.}
    \label{fig:analysis2_addi_quan}
\end{figure}

We measure object-related and color-related CLIP similarity following the same evaluation protocol used for the $k=1$ experiment in the main paper. As shown in \figref{fig:analysis2_addi_quan}, color-related similarity sharply increases at $k=1$ and saturates thereafter, demonstrating that the dominant singular direction primarily captures style. In contrast, object-related similarity increases gradually as $k$ grows, indicating that higher-rank components encode structural information.

\begin{figure*}[t]
    \centering
    \includegraphics[width=.9\linewidth]{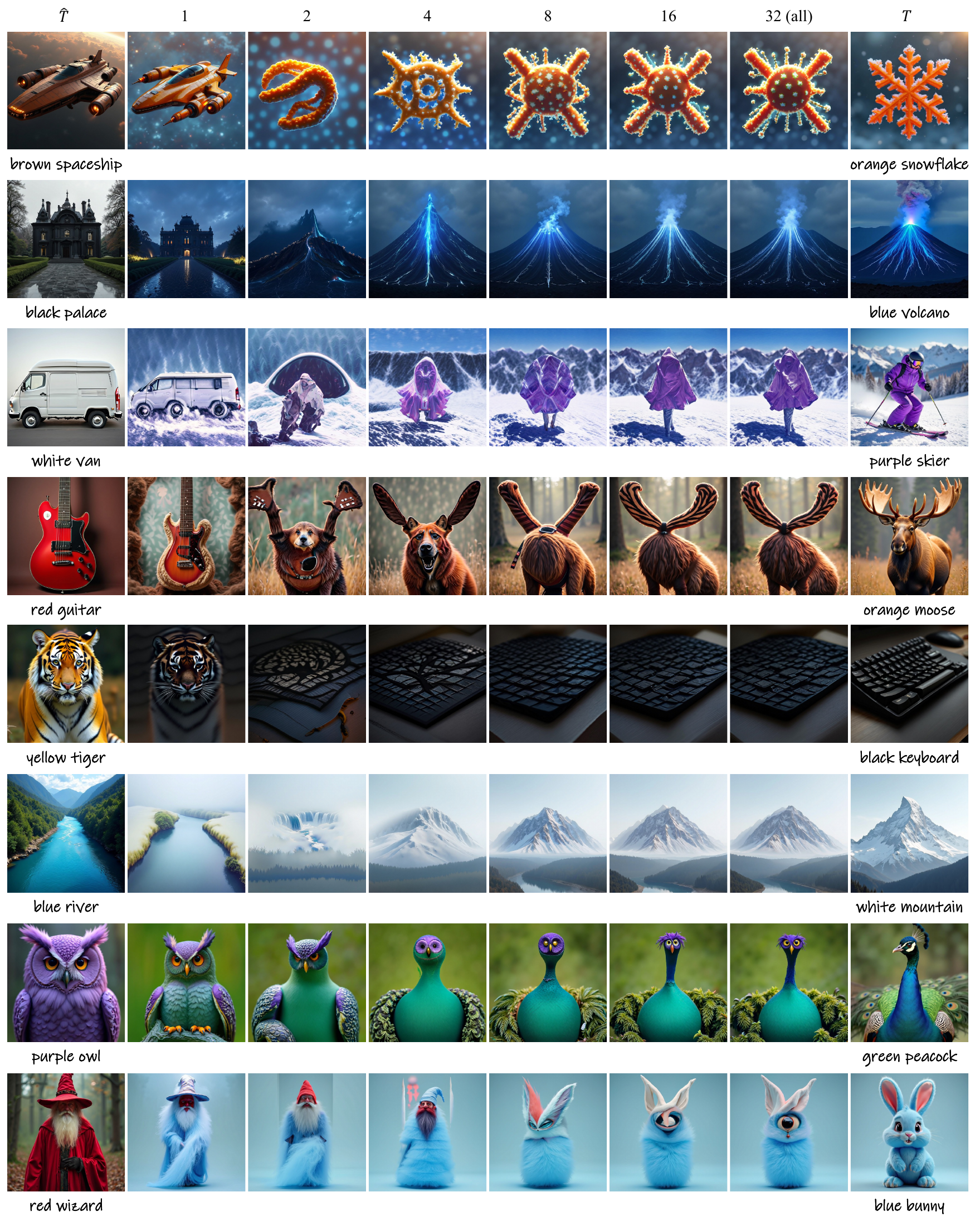}
    \caption{Qualitative results of SVD-guided feature replacement with varying $k$. From left to right: the baseline output generated from $\hat{T}$, SVD-guided outputs with $k \in \{1,2,4,8,16,32\}$, and the baseline output generated from $T$.}
    \label{fig:analysis2_addi_qual}
\end{figure*}

Qualitative examples in \figref{fig:analysis2_addi_qual} show a similar trend: the $k=1$ output transfers texture and color while preserving object shape, whereas larger $k$ values begin to alter geometry and object identity. These results further support our main finding that the first principal component of $F_3$ predominantly encodes style, also justifying our exponential reweighting design in the main method.

\subsection{Analysis of the exponential decay rate}
\begin{figure}[t]
    \centering
    \includegraphics[width=.9\linewidth]{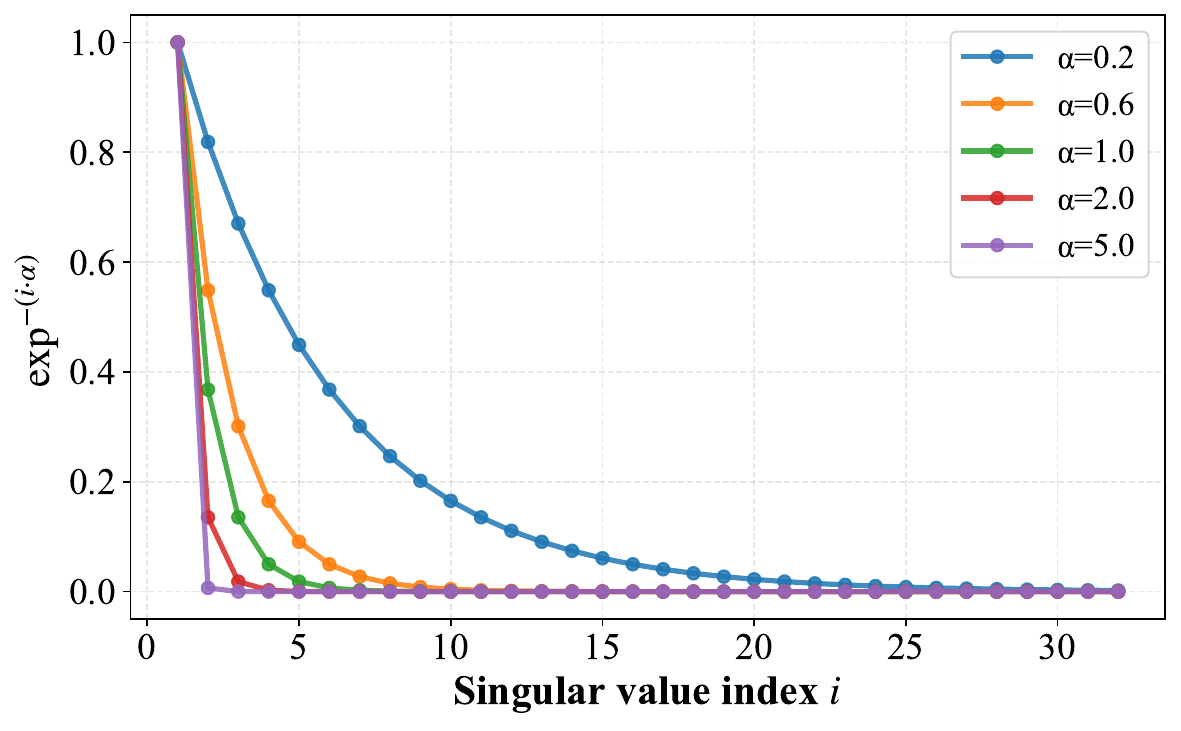}
    \caption{Visualization of exponential decay rates $\alpha$ with respect to the singular value index $i \in \{0, 1, \dots, r-1\}$.}
    \label{fig:ablation_alpha_graph}
\end{figure}

We further conduct an additional ablation study on the exponential decay rate $\alpha$ in Principal Feature Blending. As shown in \tabref{tab:ablation_alpha}, our method remains robust across different values of $\alpha$, exhibiting only a minor trade-off between style fidelity and prompt fidelity. \figref{fig:ablation_alpha_graph} provides a visualization of how varying $\alpha$ controls the exponential decay of weights across singular values. 

Decreasing $\alpha$, which increases the influence of the higher-rank singular components, naturally elevates the risk of content leakage during style injection, resulting in a decrease in prompt fidelity. This behavior is consistent with our hypothesis that the dominant singular value predominantly encodes style-related information over the remaining components.
We set $\alpha=1.0$ as it provides the most balanced performance.

\begin{table}[h]
    \caption{
    Additional ablation study on exponential decay rate ($\alpha$) in Principal Feature Blending (PFB). The symbol $\uparrow$ indicates that higher is better. The best and second-best results are highlighted in \textbf{bold} and \underline{underline}, respectively.
    }
    \centering
    \footnotesize
    \begin{tabular}{c|ccc} 
    \toprule
      alpha ($\alpha$)  
      & $S_\text{txt}$ $\uparrow$ 
      & $S_\text{img}$ $\uparrow$ 
      & $S_\text{harmonic}$ $\uparrow$ \\
    \midrule
     0.2 & 0.323 & \textbf{0.640} & 0.429 \\
     0.6 & 0.331 & \underline{0.631} & 0.434 \\
     1.0 (ours) & \underline{0.334} & 0.630 & \textbf{0.437} \\
     2.0 & \underline{0.334} & 0.624 & \underline{0.435} \\
     5.0 & \textbf{0.335} & 0.621 & \underline{0.435} \\
    \bottomrule
    \end{tabular}
    \label{tab:ablation_alpha}
\end{table}

\section{Details of the dual-stream generation mechanism}

We provide a detailed description of our dual-stream generation process in Algorithm~\ref{alg:ours}. 
Both the \textit{content path} and \textit{generation path} are conditioned on the same text prompt $T$ (``\texttt{<content> in <style>}'') and are executed jointly within a single inference batch. 
Using identical conditioning prevents semantic mismatch between the two streams and ensures that both evolve under the same textual supervision.

The content path follows the original inference process of the pre-trained model without modification, producing a sequence of features $\{F_s^{\text{con}}\}_{s=1}^S$, which serve as a structural reference. 
Meanwhile, the generation path produces its own feature sequence $\{F_s^{\text{gen}}\}_{s=1}^S$, which is selectively modulated by our proposed mechanisms (PFB, SAC). 
Throughout inference, the content path provides structural guidance to the generation path, enabling it to preserve spatial consistency while integrating style information from the reference style image.

\begin{algorithm}[t]
\caption{Dual-path style-personalized image generation}
\label{alg:ours}
\textbf{Input}: Style reference image $I^{\mathrm{sty}}$, text prompt $T$ \\ 
\textbf{Output}: Stylized image $I^{\mathrm{gen}}$

\begin{algorithmic}[1]

\STATE $\{F_s^{\mathrm{sty}}\}_{s=1}^{S} \leftarrow \mathcal{E}_I(I^{\mathrm{sty}})$
\textcolor{blue}{\small \# Multi-scale style features}

\STATE Initialize $F_0^{\mathrm{con}}, F_0^{\mathrm{gen}}$  
\textcolor{blue}{\small \# Same initial condition, same prompt $T$}

\FOR{$s = 1$ to $S$}

    \STATE \textcolor{blue}{\small\# (1) Dual-stream iterative update (Eq.~(1), (2))}
    \STATE $F_s^{\mathrm{con}} \leftarrow \mathcal{M}(F_{s-1}^{\mathrm{con}}, \mathcal{E}_T(T))$
    \STATE $F_s^{\mathrm{gen}} \leftarrow \mathcal{M}(F_{s-1}^{\mathrm{gen}}, \mathcal{E}_T(T))$

    \IF{$s = 3$}
        \STATE \textcolor{blue}{\small\# (2) Principal Feature Blending (PFB)}
        \STATE $F_3^{\mathrm{gen}} \leftarrow 
            \Phi(F_3^{\mathrm{sty}}) + (F_3^{\mathrm{gen}} - \Phi(F_3^{\mathrm{gen}}))$
    \ENDIF

    \IF{$s \in \mathbf{S}_{\mathrm{fine}}$}
        \STATE \textcolor{blue}{\small\# (3) Structural Attention Correction (SAC)}
        \STATE $Q_s^{\mathrm{gen}} \leftarrow Q_s^{\mathrm{con}} = W_Q F_s^{\mathrm{con}}$
        \STATE $K_s^{\mathrm{gen}} \leftarrow K_s^{\mathrm{con}} = W_K F_s^{\mathrm{con}}$
    \ENDIF

\ENDFOR

\STATE $I^{\mathrm{gen}} \leftarrow \text{Decoder}(F_S^{\mathrm{gen}})$
\STATE \textbf{return} $I^{\mathrm{gen}}$

\end{algorithmic}
\end{algorithm}

\section{Implementation details}

\subsection{Implementation setup of comparison models}

We conduct extensive comparisons against existing style-personalized image generation methods. To ensure fair and reproducible evaluation, all baseline models are run using publicly released implementations and their default hyperparameters, without additional tuning or prompt engineering unless explicitly required. We categorize baselines into two groups:
(1) tuning-based approaches, which require style-specific fine-tuning before inference, and
(2) training-free or pre-trained approaches, which operate directly without per-style optimization.
For each method, we follow the official configurations provided in their respective repositories unless otherwise stated.

\paragraph{Tuning-based approaches}
These methods require fine-tuning a model for each reference style image. For each style reference, we performed style-specific fine-tuning following the official instructions of each repository, and report the \textbf{total runtime} consisting of both (1) training time per style and (2) inference time per image in the main paper.

\begin{itemize}
    \item \textbf{B-LoRA}~\cite{frenkel2024implicitstylecontentseparationusing}:  
    Official implementation: \url{https://github.com/yardenfren1996/B-LoRA}
    
    \item \textbf{DB-LoRA}~\cite{ryu2023low}:  
    Official implementation: \url{https://github.com/huggingface/diffusers/tree/main/examples/dreambooth}

    \item \textbf{DreamStyler}~\cite{ahn2024dreamstyler}:  
    Official implementation: \url{https://github.com/webtoon/dreamstyler}

    \item \textbf{StyleDrop}~\cite{sohn2023styledrop}:  
    Unofficial PyTorch reproduction: \url{https://github.com/zideliu/StyleDrop-PyTorch}
\end{itemize}

\paragraph{Training-free or pre-trained approaches}
These methods do not require additional fine-tuning per style. Instead, they operate either using a pre-trained style adapter or through direct inference-time conditioning. We evaluate all methods using their \textbf{official inference settings} and do not perform retraining or additional dataset-specific tuning.

\begin{itemize}
    \item \textbf{IP-Adapter}~\cite{ye2023ipadaptertextcompatibleimage}:  
    Official implementation: \url{https://github.com/tencent-ailab/IP-Adapter}

    \item \textbf{StyleAligned}~\cite{hertz2024stylealignedimagegeneration}:  
    Official implementation: \url{https://github.com/google/style-aligned}

    \item \textbf{CSGO}~\cite{xing2024csgo}:  
    Official implementation: \url{https://github.com/instantX-research/CSGO}

    \item \textbf{StyleAR}~\cite{wu2025stylear}:  
    Official implementation: \url{https://github.com/wuyi2020/StyleAR}
\end{itemize}

All models are evaluated under a unified hardware environment using a single NVIDIA A6000 GPU with PyTorch.

\subsection{Styles and prompts for generation}
\begin{figure*}[h]
    \centering
    \includegraphics[width=0.90\linewidth]{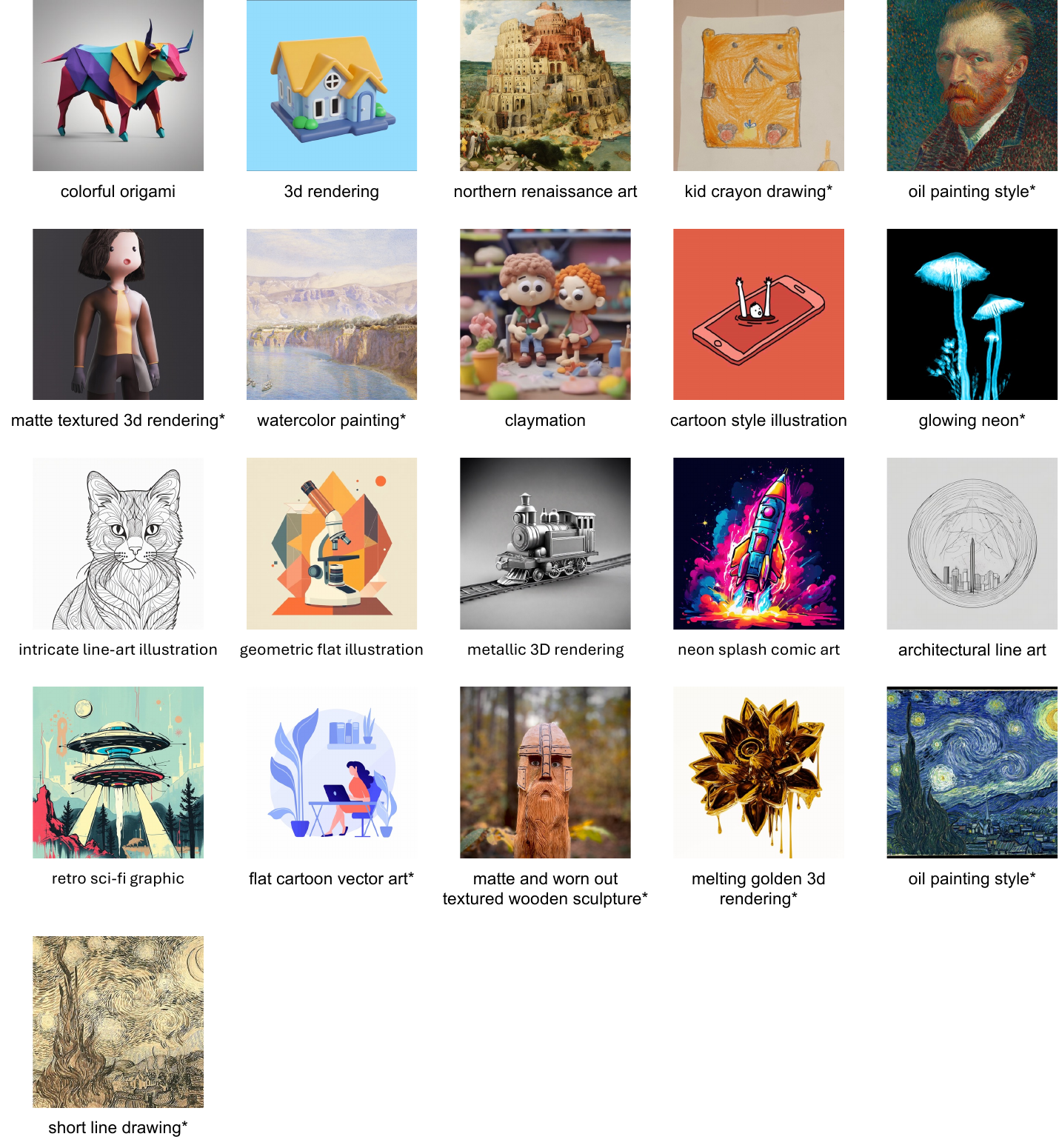}
    \caption{Style images and corresponding prompts. The symbol * indicates those used for quantitative evaluation.}
    \label{fig:style_prompts}
\end{figure*}

\figref{fig:style_prompts} presents the style prompts for the reference images used in the paper. Images marked with * indicate those used for the quantitative evaluation, for which we use the same prompts as Finestyle \cite{zhang2024finestyle}.
The style prompts serve as a high-level guide during the generation process, allowing the model to better align visual features with the target style. This provides a lightweight, training-free alternative to methods that require additional training. Note that our method does not rely on detailed prompts; simple, high-level style categories (e.g., ``oil painting," ``3d rendering," etc.) are sufficient.

\subsection{User Study Details}

To complement our quantitative evaluation, we conduct a user study involving 30 participants (ages 20s--50s). Participants compare results across two criteria: \textbf{prompt fidelity} (semantic alignment with text) and \textbf{style fidelity} (visual similarity to the reference style). Each comparison presents participants with a reference style image, a target text prompt, and outputs from multiple models.

We select comparison models based on their quantitative performance: StyleAligned~\cite{hertz2024stylealignedimagegeneration} and IP-Adapter~\cite{ye2023ipadaptertextcompatibleimage}, which achieved the highest $S_{\text{img}}$ (style fidelity), and DB-LoRA~\cite{ryu2023low} and B-LoRA~\cite{frenkel2024implicitstylecontentseparationusing}, which achieved the highest $S_{\text{txt}}$ (prompt fidelity). This selection ensures that the user study compares the strongest-performing baselines under each metric.

\begin{figure*}[t]
    \centering
    \includegraphics[width=1\linewidth]{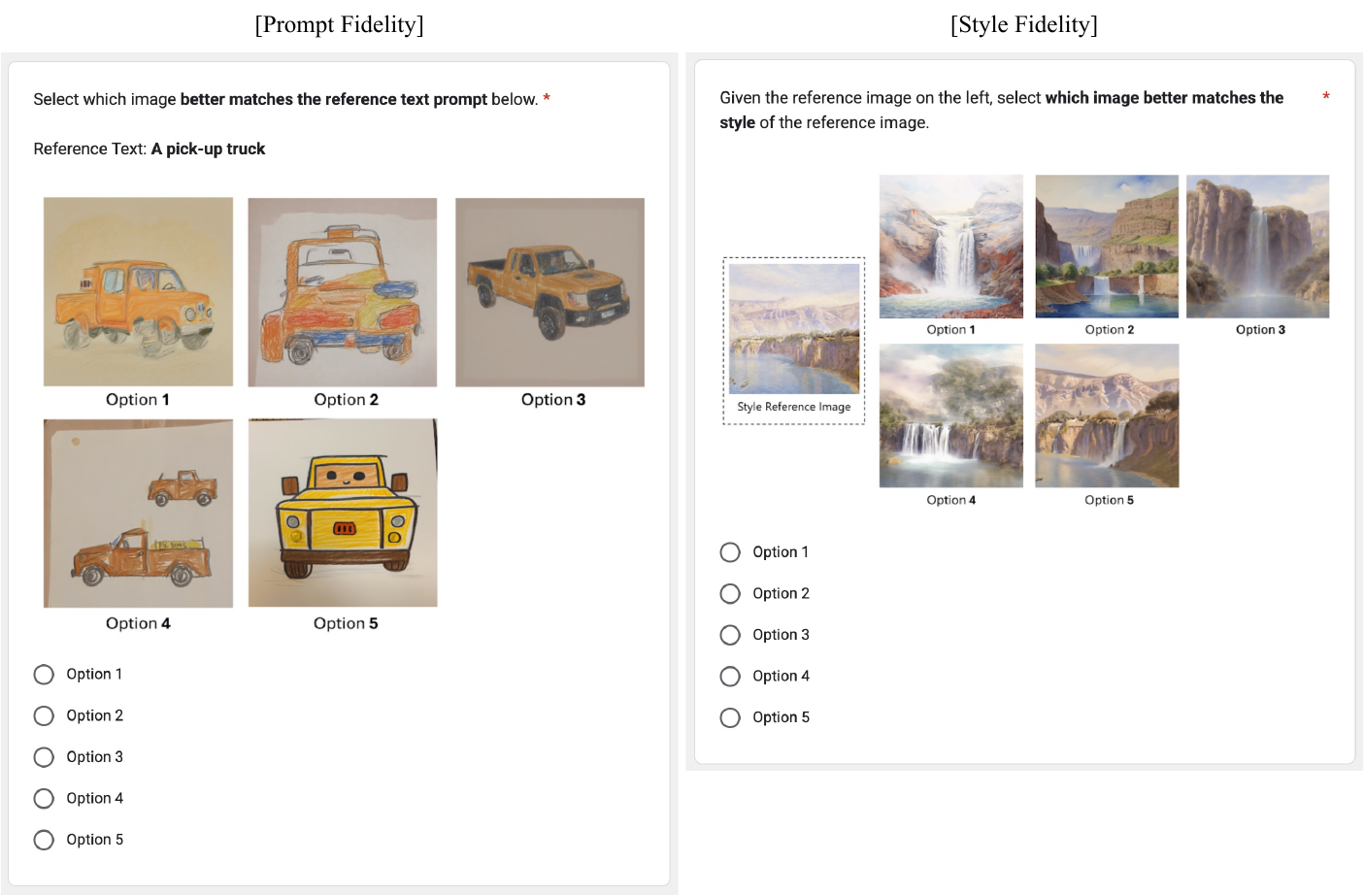}
    \caption{Example interface used in the user study. Participants selected the best-performing method among five candidates for each evaluation criterion.}
    \label{fig:userstudy_ui}
\end{figure*}

As shown in \tabref{tab:userstudy}, our method achieves the highest preference in \textbf{prompt fidelity} (35.3\%) while maintaining competitive \textbf{style fidelity} (32.0\%). Notably, prompt-tuned baselines (DB-LoRA, B-LoRA) exhibit strong semantic alignment but fail to preserve style, while style-focused baselines (StyleAligned, IP-Adapter) preserve style but lack semantic consistency. An example of the interface used in the study is shown in \figref{fig:userstudy_ui}.

\begin{table}[h]
\centering
\caption{User study preference results (percentage).}
\label{tab:userstudy}
\small
\begin{tabular}{lcc}
\toprule
\textbf{Model} & \textbf{Prompt Fidelity $\uparrow$} & \textbf{Style Fidelity $\uparrow$} \\
\midrule
StyleAligned~\cite{hertz2024stylealignedimagegeneration} & 4.3\% & 30.7\% \\
IP-Adapter~\cite{ye2023ipadaptertextcompatibleimage} & 5.0\% & 23.3\% \\
DB-LoRA~\cite{ryu2023low} & 26.7\% & 8.3\% \\
B-LoRA~\cite{frenkel2024implicitstylecontentseparationusing} & 28.7\% & 5.7\% \\
\midrule
Ours & \textbf{35.3\%} & \textbf{32.0}\% \\
\bottomrule
\end{tabular}
\end{table}

\section{Incorporating ours into other scale-wise autoregressive models} 

Our method is designed to be model-agnostic within the family of scale-wise autoregressive generative models, as it operates directly without modifying model weights or requiring retraining. To validate its generalization ability, we apply our method to two additional models beyond our primary backbone, Infinity-2B \cite{han2025infinity}.

\begin{figure*}[t]
    \centering
    \includegraphics[width=.9\linewidth]{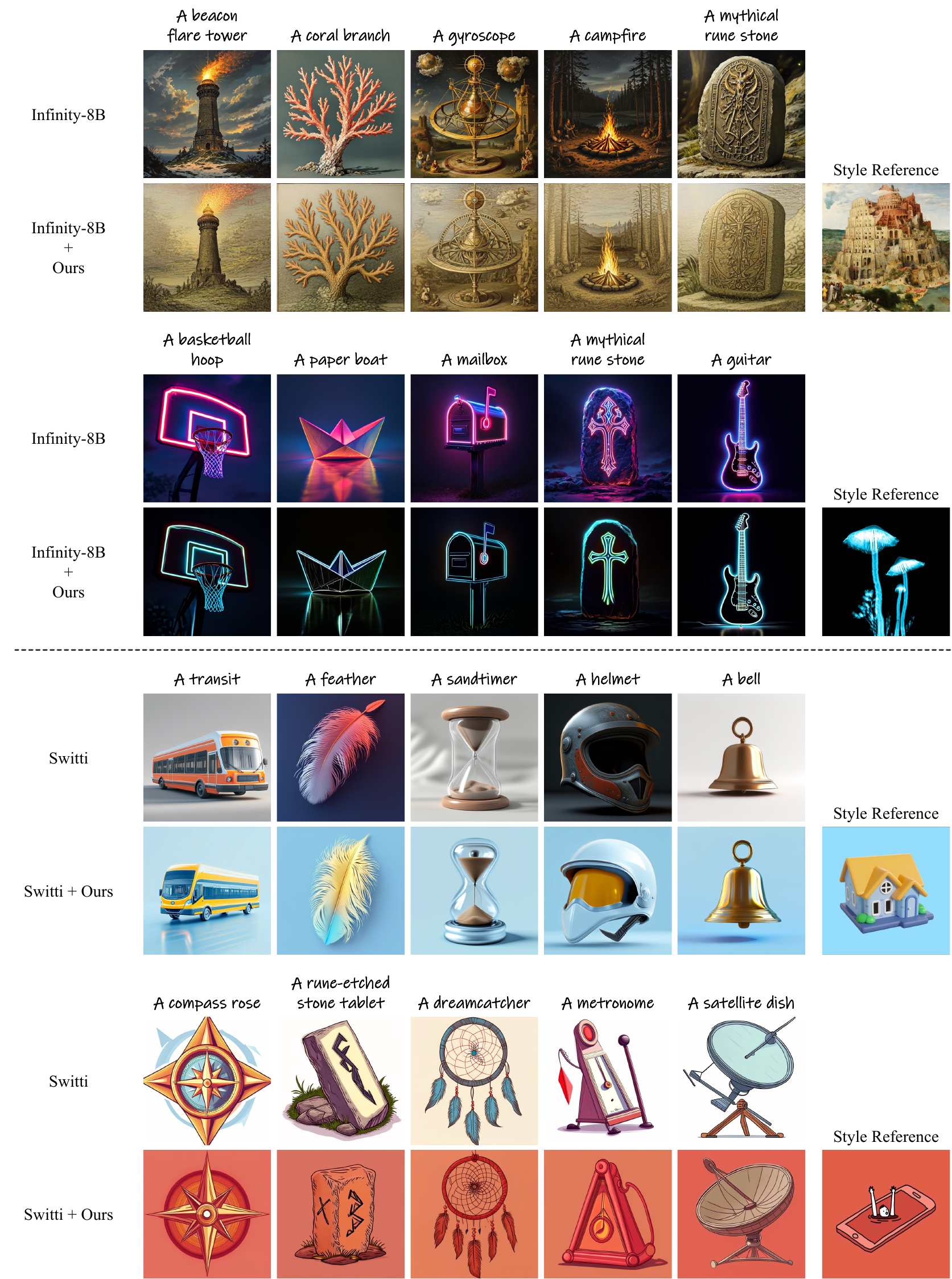}
    \caption{Qualitative results of applying our method to other scale-wise autoregressive models.}
    \label{fig:general}
\end{figure*}

We first implement our method on Infinity-8B, a larger variant of our baseline model with increased capacity and parameter count. As shown in \figref{fig:general}-(Top), our method produces consistent and stable stylization effects across this stronger model configuration, demonstrating robustness to architectural scaling without additional tuning or adaptation. We further apply our method to Switti \cite{voronov2024switti}, a distinct scale-wise autoregressive text-to-image model that differs structurally from Infinity. Despite architectural differences, our plug-and-play modules function reliably without modification, producing coherent, style-personalized generations, as shown in \figref{fig:general}-(Bottom). This result supports our approach to generalizing across models that share the scale-wise autoregressive generation paradigm.

\section{Future work and limitations}

Our work presents a training-free style-personalized image generation framework grounded in a comprehensive analysis of a scale-wise autoregressive model. By identifying a key step that significantly influences the output image and demonstrating that dominant singular components of its feature space effectively capture style information, we establish a principled mechanism for style extraction and injection. We believe that this analysis opens up several promising future directions, enabling more precise and flexible control over style, content, and other visual attributes in personalized image generation systems.

Despite these strengths, our method faces limitations when the style reference image contains heterogeneous or conflicting stylistic attributes (e.g., mixed artistic media or multiple visual motifs), as it lacks an explicit mechanism to disentangle and selectively transfer specific sub-styles. Since our style extraction relies on dominant singular components, the injected style may reflect a blended representation of multiple styles rather than a feature representing a single, isolated style. Future research could incorporate localized style decomposition, spatially variant basis representations, or user-guided selection to enable more fine-grained style control.

\section{Additional qualitative results}

\subsection{Additional results}
\begin{figure*}[t]
    \centering
    \includegraphics[width=1\linewidth]{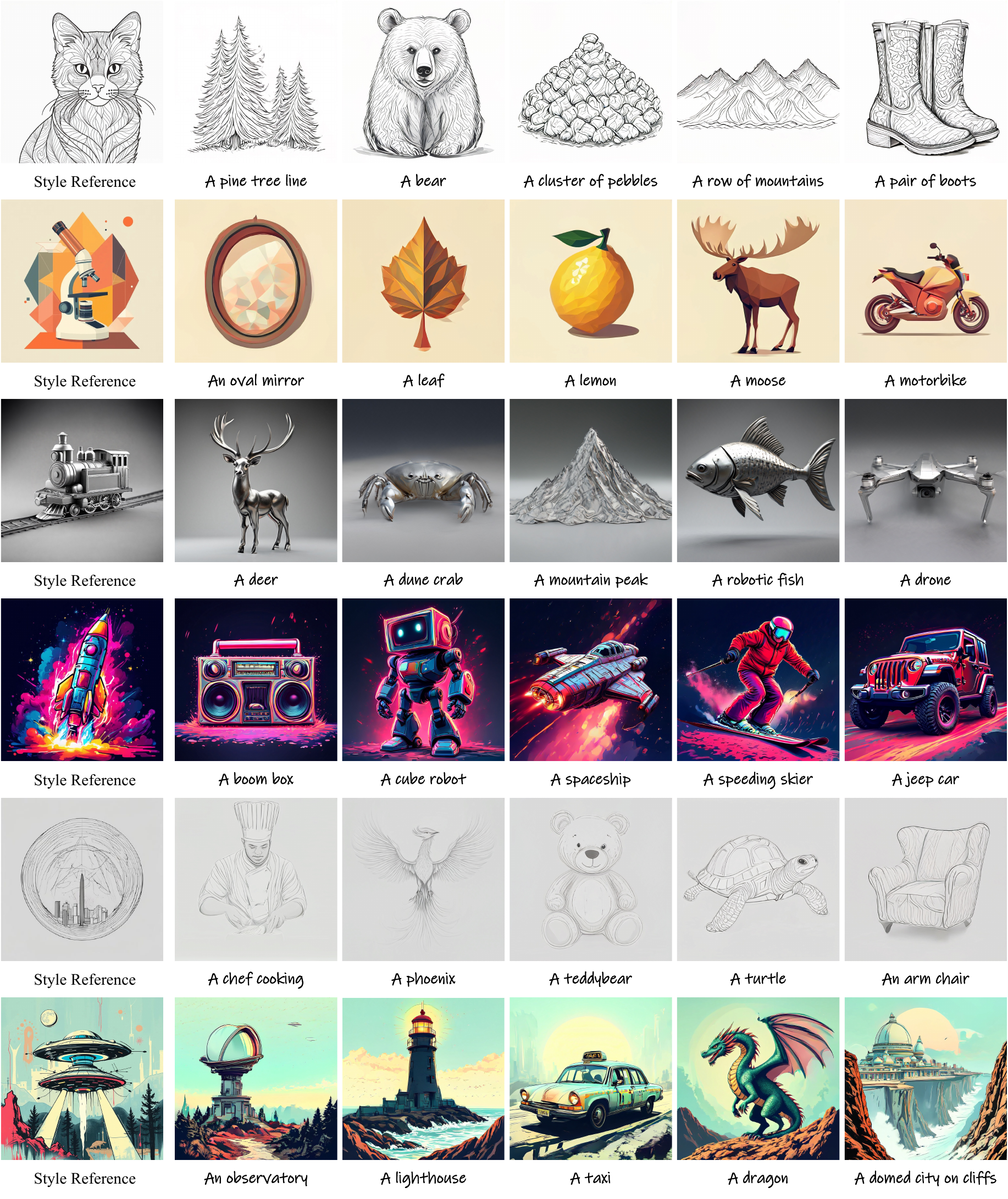}
    \caption{Various style-personalized results of our model.}
    \label{fig:add_qualitative}
\end{figure*}

\figref{fig:add_qualitative} presents additional qualitative results demonstrating that our method faithfully transfers style-specific information from the reference image while suppressing irrelevant details, effectively avoiding content leakage or mode collapse. This enables expressive and robust style personalization that generalizes well across diverse scenes and artistic styles.
\subsection{Style-aligned image generation}
\begin{figure*}[t]
    \centering
    \includegraphics[width=0.55\linewidth]{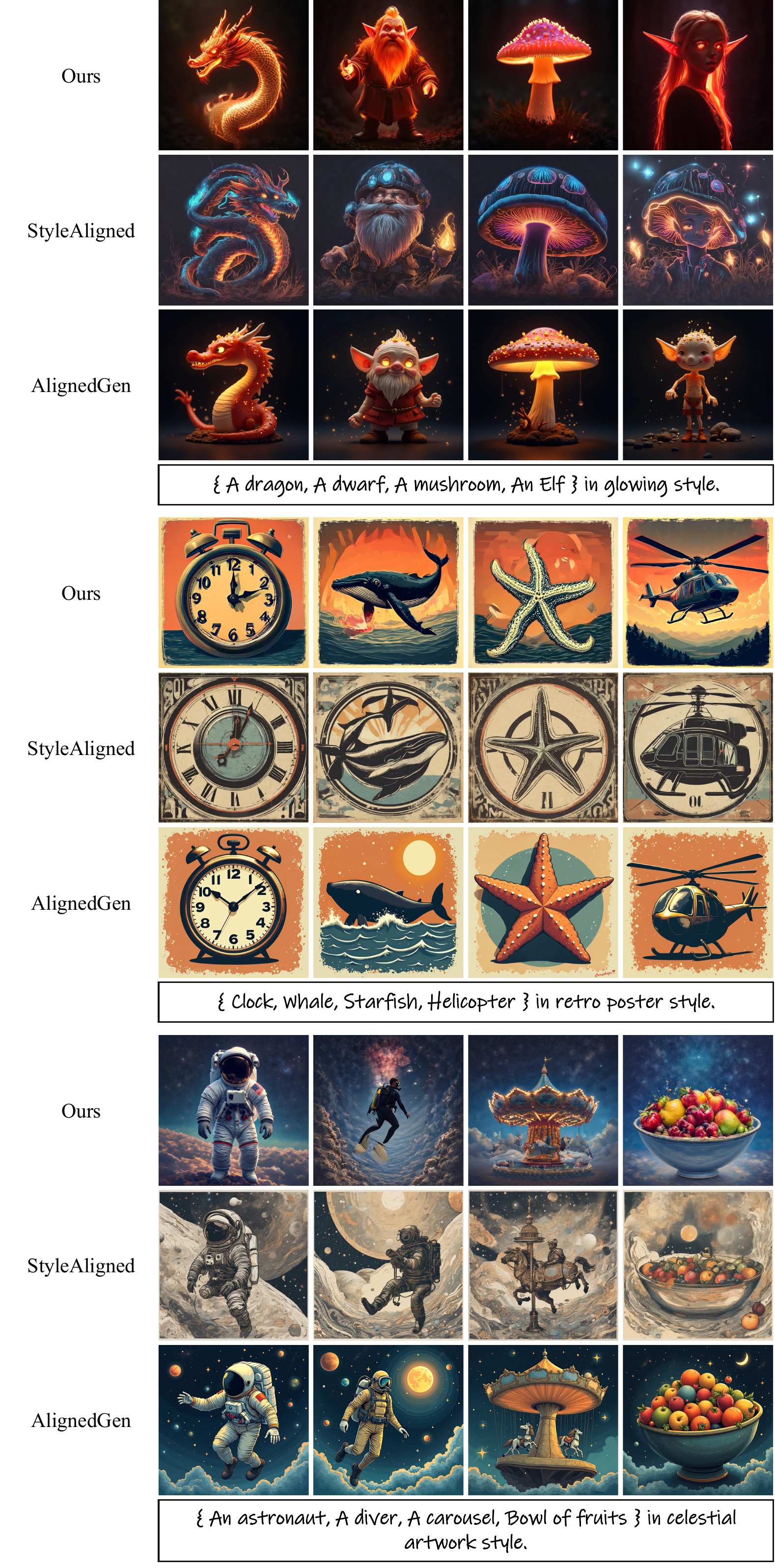}
    \caption{Style-aligned image generation results with text-only style descriptions. Each row represents a different content prompt, and each column applies a distinct style, as described in the text.}
    \label{fig:style_aligned}
\end{figure*}

Furthermore, we demonstrate that our model can perform style-aligned image generation using only a style prompt, without requiring a reference style image, by including a dedicated style pathway in the same batch derived from the style text prompt and leveraging its third feature as the style representation.
As shown in \figref{fig:style_aligned}, our model shows competitive performance compared to representative style-aligned image generation models \cite{hertz2024stylealignedimagegeneration,zhang2025alignedgen}, indicating its capability in style-aligned image generation.
These results validate that our method can operate effectively in both image-guided and text-guided style-related generation scenarios in a unified and training-free manner.


\clearpage
\clearpage
{
    \small
    \bibliographystyle{ieeenat_fullname}
    \bibliography{main}
}

\end{document}